\documentclass[10pt,twocolumn,letterpaper]{article}

\usepackage[pagenumbers]{cvpr} 
\usepackage{tabularx}
\usepackage{makecell}
\usepackage{fontawesome5}
\usepackage{pifont}
\usepackage[table]{xcolor}
\usepackage{multirow}

\definecolor{tableblue}{RGB}{209,226,252} 
\definecolor{lighttableblue}{RGB}{212, 236, 252}

\definecolor{cvprblue}{rgb}{0.21,0.49,0.74}
\usepackage[pagebackref,breaklinks,colorlinks,allcolors=cvprblue]{hyperref}
\usepackage{amssymb}
\usepackage{bbold}

\def\confName{CVPR}
\def\confYear{2027}

\title{RICOLA: \underline{Ri}gid \underline{Co}mponent Discovery without \underline{La}bels from 3D Trajectories}

\author{Xiaojie Zhang$^{1}$
\and
Tom Fischer$^{1}$
\and
Viktor Larsson$^{2}$
\and
Lukas Rosenberger Schmid$^{1}$ 
\and
\\$^{1}$University of Technology Nuremberg UTN \quad $^{2}$Lund University
}

\begin{document}
\maketitle
\begin{abstract}
Understanding dynamic scenes goes beyond observing how objects move to reasoning about how they are physically organized. Rigid components offer a compact representation of scene structure through 3D rotations and translations, while 3D point trajectories provide motion cues for discovering them. However, noisy trajectories make motion relationships ambiguous, and rigidity consistency alone cannot prevent unnecessary fragmentation. We propose a self-supervised framework that jointly learns trajectory representations and rigid-component assignments by fusing motion and appearance features with competitive slot attention. A novel motion-aware redundancy regularizer complements geometric rigidity constraints to encourage compact decompositions while preserving distinct rigid motions. Without ground-truth component annotations or rigid poses, our method discovers temporally consistent rigid components across rigid and articulated scenes. Experiments on Kubric and iTACO outperform current unsupervised methods and achieve competitive performance to supervised methods across different trajectory sources.
\end{abstract}    
\section{Introduction}
\label{sec:introduction}

Understanding how a scene is organized into coherent moving components is essential for a range of applications, including robotic manipulation~\cite{byravan2017se3,bharadhwaj2024track2act,qian2026motionbits,werby2025articulated,kerr2024robot}, autonomous driving~\cite{yang2021learning,lin2025voteflow}, articulated object understanding~\cite{le2025articulate,li2026particulate,yu2026part,peng2026itaco,Delitzas_2026_CVPR}, dynamic scene modeling~\cite{mazur20264d,shapeofmotion, schmid2024khronos}, and future motion prediction~\cite{Lee_2026_CVPR,lei2026world}. Rigid components offer a compact representation of this structure, describing the motion of independent objects and articulated parts through shared 3D rotations and translations.

Complete rigid decomposition of a scene requires observing its motion over time. A pair of scissors, for example, may move as one unit while closed and reveal two distinct rigid parts when opened.
The motion of individual points in the scene can therefore provide useful cues. Recent advances in 3D point tracking~\cite{ngo2025delta,zhang2026tapip3d,xiao2024spatialtracker,mose3,sm4rt,jain2026trackeverything} have made it increasingly practical to capture such motion through 3D point trajectories, providing a low-level representation for discovering rigid components.

\begin{figure}[t]
    \centering
    \includegraphics[width=\linewidth]{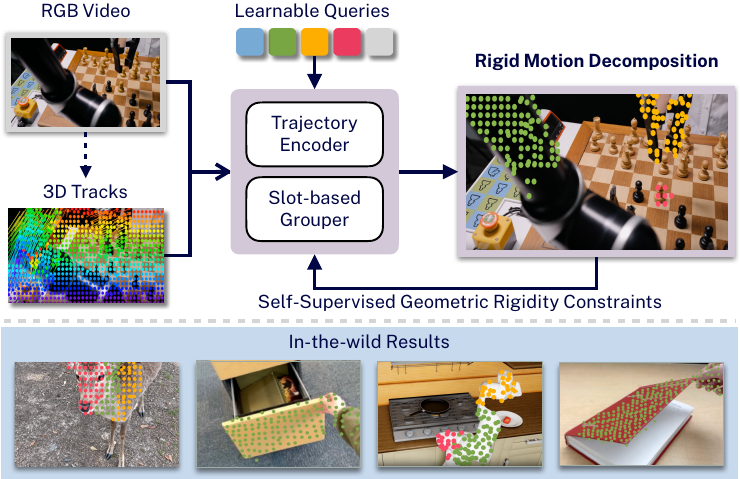}
    \caption{We propose an end-to-end framework for rigid motion decomposition by grouping 3D tracks into an unknown number of individual components via self-supervised geometric rigidity constraints balanced by slot redundancy regularization.}
    \label{fig:teaser}
\end{figure}
There are two main challenges for this task. First, decomposition relying on motion alone is inadequate as the motion cues are often ambiguous under noise. Second, determining the appropriate number of components is challenging, as rigidity consistency alone cannot prevent unnecessary fragmentation. A single rigid component may be fragmented into multiple groups that remain individually rigid, while distinct components with similar motions may be incorrectly merged. 


Existing approaches typically address rigid motion decomposition through geometric model fitting~\cite{barath2019progressive} or learned structured motion representations~\cite{sm4rt,mose3}. Geometric fitting methods can be sensitive to trajectory estimation errors and require careful selection of inlier thresholds across different track sources. Concurrent learning-based approaches~\cite{sm4rt,mose3} rely on ground-truth rigid motion annotations to supervise structured motion representation. 


To address these challenges, we propose RICOLA, a self-supervised framework that jointly learns trajectory representations and a slot-based grouper from 3D point trajectories (see ~\cref{fig:teaser}). We use visual appearance as complementary evidence to resolve ambiguous motion cues and leverage slot attention~\cite{slotattention} to discover candidate rigid components. We train the framework using geometric rigidity constraints to encourage rigidity within components, together with redundancy regularization to penalize unnecessary fragmentation. At inference, our framework can be combined with a wide range of 3D point trackers and 3D reconstruction models to decompose rigid motions for casual videos. 



In summary, our main contributions are:
\begin{itemize}
    \item We propose a self-supervised framework for rigid body discovery that combines 3D motion and appearance in trajectory-level representations to produce temporally consistent component assignments without ground-truth component labels.

    \item We introduce a motion-aware redundancy regularizer that complements pairwise rigidity by discouraging unnecessary splits of components with compatible motions while preserving components with distinct motions.

    \item Experiments on Kubric and iTACO outperform current unsupervised methods and achieve competitive performance against supervised baselines for rigid object and part discovery. Our code will be released open source.
\end{itemize}

\section{Related Work}
We summarize related methods for motion-based segmentation in~\cref{tab:baseline_comparison}.

\subsection{Unsupervised Object Discovery}

Unsupervised object discovery~\cite{simeoni2021localizing,Wang_2022_CVPR,fan2024adaptive,ouyang2026qasa,liu2026metaslot,kipf2022conditional,NEURIPS2022_ba1a6ba0,slotcontrast,nguyen2026tsa} aims to identify coherent objects without ground-truth segmentation. 
Slot Attention~\cite{slotattention} provides a framework for it by representing a scene as a set of latent slots that compete to explain input features and are iteratively refined through attention. A central challenge is determining the appropriate number of active slots. Conventional slot attention uses a fixed number of slots, potentially leading to over-segmentation or insufficient grouping capacity. Recent methods ~\cite{fan2024adaptive, liu2026metaslot, ouyang2026qasa, nguyen2026tsa} address this challenge through adaptive slot selection or redundancy reduction. However, directly penalizing the number of active slots~\cite{fan2024adaptive} can conflict with reconstruction objectives, which leads to degenerate groupings. 
In contrast, we address the challenge of redundant slots with a motion-aware redundancy regularizer that guides slot-based grouping using SE(3) structure.

\begin{table}[t]
\centering

\setlength{\tabcolsep}{4pt}
\renewcommand{\arraystretch}{1.12}

\resizebox{\columnwidth}{!}{%
\begin{tabular}{lccc}
\toprule
\textbf{Method} &
\textbf{Input} &
\makecell{\textbf{Task}\\\textbf{Supervision}} &
\makecell{\textbf{Component}\\\textbf{Extraction}} \\
\midrule
\multicolumn{4}{l}{
\textbf{\textit{Classical Motion Grouping (Optimization)}}} \\[1pt]

Progressive-X~\cite{barath2019progressive}
& 3D tracks
&\ding{55}
& multi-model fitting \\

Trajectory Clustering~\cite{shi2013robust}
& 2D tracks
&\ding{55}
& spectral clustering \\

\midrule

\multicolumn{4}{l}{
\textbf{\textit{Moving Object Segmentation (Learned)}}} \\[1pt]

SegAnyMo~\cite{seganymo}
& RGB, 2D tracks, depth
& \ding{51}
& SAM2 prompting\\

GeoMotion~\cite{geomotion}
& RGB, 2D tracks
& \ding{51}
& mask pred. + SAM2 \\

GMOS~\cite{gmos}
& RGB
& \ding{51}
& mask prediction \\

\midrule

\multicolumn{4}{l}{
\textbf{\textit{Structured SE(3) Motion Modeling (Learned)}}}\\[1pt]

SM4RT~\cite{sm4rt}
& RGB
& \ding{51}
& HDBSCAN$^\ast$ \\

MoSE3~\cite{mose3}
& RGB
& \ding{51}
& HDBSCAN$^\ast$ \\

\midrule
\multicolumn{4}{l}{
\textbf{\textit{Rigid Body Discovery}}} \\[1pt]

MotionBits~\cite{qian2026motionbits} (\textit{Optimization})
& RGB, 2D tracks
&\ding{55}
& graph \\

\rowcolor{tableblue}
{Ours} (\textit{Learned})
& {RGB, 3D tracks}
& {\ding{55}}
& {learned grouping} \\

\bottomrule
\end{tabular}%
}
\caption{\textbf{Comparison of motion-based segmentation and rigid motion modeling methods.} Task supervision: requires component labels during training. Component extraction describes how discrete motion groups or rigid components are obtained. $^\ast$ Discrete rigid-part labels require additional clustering. 
}
\label{tab:baseline_comparison}

\end{table}

\subsection{Motion Segmentation}

Motion segmentation groups pixels or trajectories according to their motion. Long-term point trajectories provide useful temporal information for identifying coherently moving regions. However, trajectories estimated from real-world videos often suffer from tracking errors, depth inaccuracies, and occlusions, making reliable motion grouping challenging. Classical trajectory-based methods~\cite{shi2013robust, isack2012energy,delong2012fast,barath2019progressive} address these difficulties through motion subspace representations, graph-based clustering, or multi-model fitting. Progressive-X~\cite{barath2019progressive} combines progressive hypothesis generation with labeling optimization to recover multiple geometric models and penalize unnecessary ones. Recent learning-based video moving object segmentation methods~\cite{seganymo,geomotion,hu2026vggt4d,gmos} combine learned visual representations with motion cues~\cite{seganymo,geomotion}, while other methods discover motion cues directly from multi-view 3D reconstruction models~\cite{wang2025vggt,wang2026pi}. However, moving object segmentation alone does not solve rigid component structure. For example, a single articulated object can contain multiple components with distinct rigid motions. This distinction motivates grouping by shared 3D rigid motion rather than object identity. Our method introduces a self-supervised rigidity constraint to learn this grouping directly from 3D trajectories.
\begin{figure*}[t]
    \centering
    \includegraphics[width=1.0\linewidth]{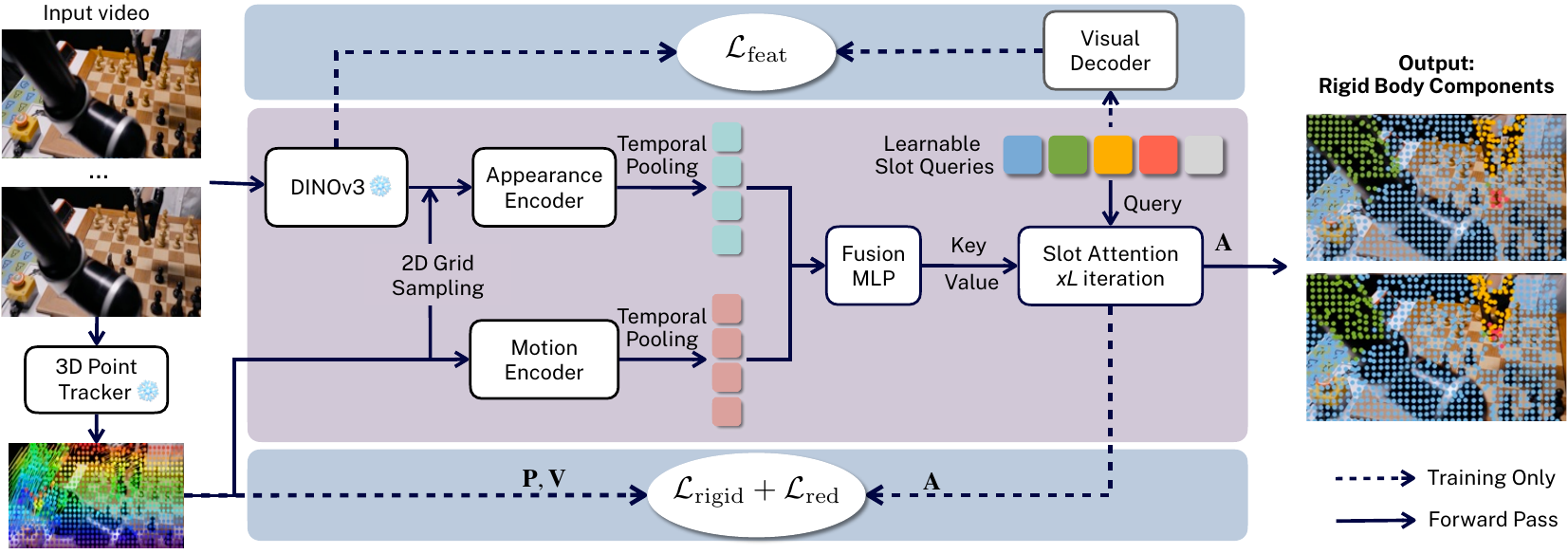}
    \caption{\textbf{Method overview.} Given an RGB video, 3D point trajectories $\mathbf{P}$ and their visibility $\mathbf{V}$, separate motion and appearance encoders extract complementary features, followed by an MLP for feature fusion. Slot queries learn to represent individual components, while slot attention iteratively groups the trajectory features into components. 
    After $T$ iterations, slot attention outputs soft trajectory-to-component assignments $\mathbf{A}$, ensuring temporally consistent grouping. The training objectives combine geometric rigidity $\mathcal{L}_{rigid}$ and redundancy regularization $\mathcal{L}_{red}$ with an auxiliary visual loss $\mathcal{L}_{feat}$. The model is trained end-to-end without ground-truth rigid component labels.  
    }
    \label{fig:method_overview}
\end{figure*}
\subsection{3D Rigid Motion Modeling and Decomposition}

Recovering rigid components in 3D requires identifying points that share a common SE(3) transform over time.  Several approaches~\cite{huang2021multibodysync,xiao2024spatialtracker,sm4rt,mose3} jointly model 3D motion and rigid structures. 
MoSE3~\cite{mose3} propose to predict dense SE(3) motion by joint learning of 3D point tracking and rigid embeddings with a differentiable pose solver supervised by \textit{SE(3) poses}. SM4RT~\cite{sm4rt} decomposes scenes into motion bases as 6D twists supervised by the ground-truth 3D point trajectories and \textit{translation of SE(3) transform}. While learned rigidity embeddings capture motion relationships, extracting discrete components requires post-hoc clustering, such as HDBSCAN~\cite{mcinnes2017hdbscan}, which may produce fragmented assignments. Rigid motion decomposition is closely related to articulated object understanding~\cite{le2025articulate,li2026particulate,yu2026part,peng2026itaco,Delitzas_2026_CVPR,ai2026articulation,li2026artpro,qu2026famos}. MotionBits~\cite{qian2026motionbits} introduces a kinematics-based definition of rigid components and an optimization-based method that groups 2D trajectories to rigid components using optical flows. These work motivate learning discrete rigid component assignments jointly with motion representations, rather than extracting them through a separate clustering stage. Our method jointly learns trajectory representations and a slot attention grouping mechanism under self-supervision, directly predicting rigid-component assignments in an end-to-end framework.

\section{Method}
\subsection{Problem Formulation}
\label{sec:problem}
Given a video $\mathcal{I}=\{I_t\}_{t=1}^{T}$, we consider $Q$ point trajectories with 3D coordinates $\mathbf{P}\in\mathbb{R}^{T\times Q\times 3}$ in the world coordinate frame, corresponding image coordinates $\mathbf{U}\in\mathbb{R}^{T\times Q\times 2}$, and visibility indicators $\mathbf{V}\in\{0,1\}^{T\times Q}$. We consider the first camera frame as the world coordinate. We denote an observation of trajectory $q$ at time $t$ by $\mathbf{p}^{t}_q$, $\mathbf{u}^{t}_q$, and $V^{t}_q$. We define a rigid part as the component unit, so a group of trajectories \textit{undergoing the same rigid motion} over time is considered as one component. Our task is to discover these rigid components avoiding redundant components.

\subsection{Overview} 
We introduce a feed-forward self-supervised framework for rigid component discovery from RGB videos and 3D trajectories (\cref{fig:method_overview}). Our architecture contains a trajectory encoder (\cref{sec:traj_encoding}) and a slot attention-based grouper (\cref{sec:slot_grouping}) supervised by the geometric rigidity constraint (\cref{sec:rigid_loss}), slot redundancy regularization (\cref{sec:red_loss}).
 
\subsection{Trajectory Representation}
\label{sec:traj_encoding}
3D trajectories provide valuable information for rigid component discovery.
However, noisy or incomplete tracks can make motion-based grouping ambiguous, particularly when the motion is small. To obtain more robust trajectory representations, we jointly encode motion and appearance cues and aggregate information across time.

\noindent\textbf{Motion encoder.}
We normalize 3D trajectory for per clip. For each 3D point, we construct the descriptor $\mathbf{d}^{t}_q=[\mathbf{p}^{t}_q;\Delta\mathbf{p}^{t}_q;\mathbf{v}^{t}_q;\|\mathbf{v}^{t}_q\|_2]\in \mathbb{R}^{10}$, where $\Delta\mathbf{p}^{t}_q$ is displacement relative to the first frame and $\mathbf{v}^{t}_q$ is velocity to the previous frame. After layer normalization and liner projections, a temporal transformer with 1D RoPE (rotary positional embeddings) attends across time within each trajectory. The visibility term $V^{t}_q$ masks invisible observations in temporal attention and mean pooling, so only visible observations contribute to the trajectory representation. This yields the trajectory-level motion representation $\mathbf{h}^{m}_q \in \mathbb{R} ^{D}$, where $D$ is the feature channel.

\noindent\textbf{Appearance encoder.}
To complement motion cues, we extract dense visual features with a frozen DINOv3~\cite{simeoni2025dinov3} backbone and sample them at the image coordinates $\mathbf{u}^{t}_q$. Temporal attention operates across observations along each trajectory, while spatial attention operates across trajectories within each frame. Together, they process the sampled features $\mathbf{f}^{t}_q$ to capture appearance consistency and relationships among tracked points. We use 4D RoPE over time and 3D position $(t,x,y,z)$, and aggregate the encoded features over visible observations to obtain the trajectory-level appearance representation $\mathbf{h}^{a}_q \in \mathbb{R} ^{D}$.

\noindent\textbf{Feature Fusion.} Finally, we combine motion $\mathbf{h}^{m}_q$ and appearance representations $\mathbf{h}^{a}_q$ into a unified trajectory embedding $\mathbf{z}_q \in \mathbb{R} ^{ D}$ using an MLP.

\subsection{Slot-based Rigid Component Discovery}
\label{sec:slot_grouping}
To discover rigid components, we learn a discrete component representation by grouping continuous trajectory embeddings $\mathbf{z}_q$ using slot attention~\cite{slotattention}. First, we initialize $K$ slot queries $\mathbf{s} \in \mathbb{R}^{K\times D}$, each representing a candidate rigid component. At each iteration, slots compete to explain parts of $\mathbf{z}_q$ through a softmax-based attention:  
\begin{equation}
\begin{aligned}
    A_{kq}=\frac{\exp(M_{kq})}{\sum_{j=1}^{K}\exp(M_{jq})},
    M_{kq}
    &=\frac{\langle\phi_{\mathrm{query}}(\mathbf{s}_k),
    \phi_{\mathrm{key}}(\mathbf{z}_q)\rangle}{\sqrt{D}}.
\end{aligned}
\end{equation}
Here, the $\phi_{query}$ and $\phi_{key}$ are learnable linear projection. The attention weights are used to aggregate trajectory features for each slot, which is iteratively updated using a gated recurrent unit (GRU). After $L$ iterations, $\mathbf{s}$ encode candidate rigid components, while the attention weights $\mathbf{A}$ define their trajectory assignments. We use $L=3$ in all experiments. At inference, we obtain hard assignment labels by taking the argmax over the slot dimension of $\mathbf{A}$. Since grouping operates on entire trajectories, component assignments remain consistent across time.

\subsection{Geometric Rigidity Constraint}
\label{sec:rigid_loss}
To constrain the rigidity without ground-truth rigid component labels, we exploit the geometric property of rigid motion that pairwise Euclidean distances between points on the same rigid body remain constant over time. This provides an effective loss without having to explicitly estimate rigid transformations through a differentiable pose solver. For two different trajectories $i$ and $j$, let $d^{t}_{ij}=\|\mathbf{p}^{t}_i-\mathbf{p}^{t}_j\|_2$. We use the mean distance over jointly visible frames:
\begin{equation}
    \bar{d}_{ij}=
    \frac{\sum_{t=1}^{T}V^{t}_iV^{t}_j d^{t}_{ij}}
    {\sum_{t=1}^{T}V^{t}_iV^{t}_j}.
\end{equation}
Only pairs with at least one jointly visible frame are included. The rigidity loss is thus
\begin{equation}
    \mathcal{L}_{\mathrm{rigid}}=
    \frac{
        \sum_t\sum_{i<j}
        V^{t}_iV^{t}_j w_{ij}(d^{t}_{ij}-\bar{d}_{ij})^2
    }{
        \sum_t\sum_{i<j}
        V^{t}_iV^{t}_j w_{ij}
    },
    \label{eq:rigidity_loss}
\end{equation}
where $w_{ij}=\sum_{k=1}^{K}A_{ki}A_{kj}$ is the co-assignment weight. This term penalizes assigning trajectories with changing relative distances to the same component.

\subsection{Slot Redundancy Regularization}
\label{sec:red_loss}
\begin{figure}
    \centering
    \includegraphics[width=\linewidth]{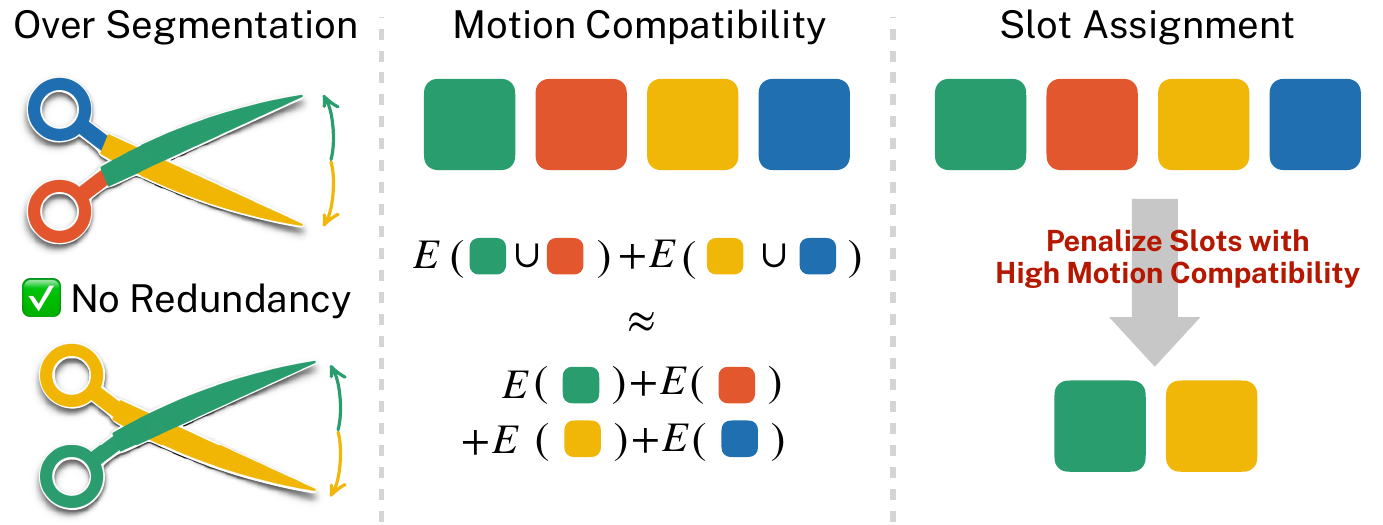}
    \caption{\textbf{Motion-aware slot redundancy regularization.} We compare the rigid alignment errors of two slots fitted separately and jointly. Slots compatible with a shared rigid motion receive a redundancy penalty that encourages their assignments to merge during training.}
    \label{fig:slot_red}
\end{figure}

The rigidity loss encourages trajectories within each slot to move rigidly, but does not prevent redundant slots. 
Since trajectories are noisy, more slots always reduce the rigidity residual and lead to unnecessarily fragmentation. Directly penalizing the number of occupied slots, however, tends to merge independently rigid components and leads to degeneration of groupings in our experiments. Inspired by energy-based multi-model fitting~\cite{isack2012energy,delong2012fast}, we encourage compact decompositions by penalizing slot pairs that share similar rigid motion (\cref{fig:slot_red}).

\noindent\textbf{Measuring motion compatibility.} To determine whether two slots describe the same rigid body, we compare how well their trajectories can be explained by a shared rigid transformation. If they belong to the same rigid body, fitting a single transformation to their union should have approximately the same error.

First, we choose the frame $r$ with the most visible trajectories as the reference. For slot $k$ and frame $t$, we define an energy function as follows:
\begin{equation}
    E_t(k)=
    \min_{\mathbf{R}\in SO(3),\,\mathbf{b}\in\mathbb{R}^{3}}
    \sum_q\omega^{t}_{kq}
    \big\|\mathbf{R}\mathbf{p}^{r}_q+\mathbf{b}-\mathbf{p}^{t}_q\big\|_2^2,
    \label{eq:slot_alignment}
\end{equation}
where $\omega^{t}_{kq}=V^{r}_qV^{t}_qA_{kq}$ combines visibility and slot assignment. We solve this weighted rigid alignment using weighted Umeyama~\cite{umeyama1991least}.  

For two slots $a$ and $b$, we measure the additional error introduced by
fitting their trajectories with a single transformation rather than
two separate transformations. Let $E_t(a\cup b)$ denote the alignment
error of their combined trajectories, and let $n^t_k=\sum_q\omega^t_{kq}$
denote the effective observation count. Over the set $\mathcal{T}_{ab}$ of
non-reference frames where both slots have sufficient observations, we define:
\begin{equation}
    \Delta_{ab}=
    \frac{
        \sum_{t\in\mathcal{T}_{ab}}
        E_t(a\cup b)-E_t(a)-E_t(b)
    }{
        \max\left(1,\sum_{t\in\mathcal{T}_{ab}}
        \min(n^{t}_a,n^{t}_b)\right)
    }.
    \label{eq:merge_cost}
\end{equation}A small $\Delta_{ab}$ indicates compatibility with a shared rigid motion. We normalize the excess error by the smaller effective observation count to account for differences in slot sizes and prevent large slots from dominating the comparison. Then, we convert the error into a soft compatibility score:
\begin{equation}
    c_{ab}=\mathbb{1}^{\mathrm{valid}}_{ab}
    \sigma\!\left(\frac{\delta-\Delta_{ab}}{\tau}\right),
    \label{eq:motion_compatibility}
\end{equation}
where $\sigma$ is the sigmoid function, $\delta$ sets the tolerated motion discrepancy, $\tau>0$ controls its softness, and $\mathbb{1}^{\mathrm{valid}}_{ab}$ indicates sufficient temporal overlap for a valid comparison. This soft comparison avoids making a hard decision based on potentially noisy trajectory measurements.

\noindent\textbf{Penalizing redundant slots.}
Given $c_{ab}$, we encourage slots with similar
motions to merge their trajectory assignments. We measure the total trajectory-to-slot assignment mass of slot $k$ as $m_k=\sum_{q=1}^{Q}A_{kq}$. A small $m_k$ means that the slot accounts for little of the observed trajectory set. We penalize the smaller mass in each compatible pair:
\begin{equation}
    \mathcal{L}_{\mathrm{red}}=
    \frac{
        \sum_{a<b}\operatorname{sg}(c_{ab})\min(m_a,m_b)
    }{
        \max(1,\sum_{a<b}\operatorname{sg}(c_{ab}))
    },
    \label{eq:redundancy_loss}
\end{equation}
where $\operatorname{sg}$ denotes stop-gradient. Penalizing the smaller mass encourages redundant slots to consolidate their assignments while preserving slots with distinct motions. Compatibility scores are detached throughout this loss, so gradients act through assignment masses not rigid alignment. 


\subsection{Training Objectives}

\noindent\textbf{Visual feature reconstruction.}
\label{sec:rec_loss}
We use visual feature reconstruction as an auxiliary objective. Following feature reconstruction in DINOSAUR~\cite{seitzer2022bridging}, we supervise slot representations with features from DINOv3. We extend reconstruction across the video by using the same slots and trajectory assignments at every time step. For each trajectory at time $t$, a Fourier embedding $\boldsymbol{\gamma}^{t}_q=\gamma(\mathbf{p}^{t}_q,t)$ encodes its 3D position and time. The decoder predicts a feature $\hat{\mathbf{f}}^{t,k}_q=\phi_{\mathrm{dec}}(\mathbf{s}_k,\boldsymbol{\gamma}^{t}_q)$ for each slot. The final prediction combines these features using the trajectory assignments $\hat{\mathbf{f}}^{t}_q=\sum_{k=1}^{K}A_{kq}\hat{\mathbf{f}}^{t,k}_q$. 
The auxiliary loss $\mathcal{L}_{\mathrm{feat}}$ is the masked mean squared error between these predictions and the sampled DINOv3 features over visible observations. 

\noindent\textbf{Overall objective.} 
Our method is trained end-to-end without ground-truth component labels. The final objective is
\begin{equation}
    \mathcal{L}=
    \lambda_{\mathrm{rigid}}\mathcal{L}_{\mathrm{rigid}}
    +\lambda_{\mathrm{red}}\mathcal{L}_{\mathrm{red}}
    +\lambda_{\mathrm{feat}}\mathcal{L}_{\mathrm{feat}}.
    \label{eq:overall_loss}
\end{equation}
We set $\lambda_{\mathrm{rigid}}=1.0$, $\lambda_{\mathrm{feat}}=0.5$, and $\lambda_{\mathrm{red}}=0.1$. These objectives encourage rigid, non-redundant components informed by visual appearance.

\noindent\textbf{Training Details.}
Our training data contains Kubric~\cite{kubric}, ArtKubric~\cite{mose3}, Synthverse~\cite{synthverse} (articulation, animals, embodied, flowers subsets) using ground-truth 3D trajectories without rigid component annotation (e.g., SE(3) transform, rigid component masks). 
We apply image, 2D trajectory and depth augmentations. We first train for 50k steps on Kubric, then train on the mixed dataset for 30k steps. Our network contains 4.7~M trainable parameters. The first stage is trained on a single H100 GPU, then the second stage is trained on 3 H100 GPUs. The total training time is around 2 days. Further details are provided in the supplementary material.

\begin{table*}[t]
\centering
\small
\setlength{\tabcolsep}{4.5pt}
\begin{tabular}{lcccc cccc cccc}
\toprule
& \multicolumn{4}{c}{\textbf{GT}}
& \multicolumn{4}{c}{\textbf{MoSE3}}
& \multicolumn{4}{c}{\textbf{TrackEverything}} \\
\cmidrule(lr){2-5}
\cmidrule(lr){6-9}
\cmidrule(lr){10-13}
\textbf{Method}
& \textbf{ARI} $\uparrow$
& \textbf{mIoU} $\uparrow$
& \textbf{F1} $\uparrow$
& \textbf{R.} $\uparrow$
& \textbf{ARI} $\uparrow$
& \textbf{mIoU} $\uparrow$
& \textbf{F1} $\uparrow$
& \textbf{R.} $\uparrow$
& \textbf{ARI} $\uparrow$
& \textbf{mIoU} $\uparrow$
& \textbf{F1} $\uparrow$
& \textbf{R.} $\uparrow$
\\
\midrule
\multicolumn{13}{l}{\textit{Supervised with Rigid Component Labels}} \\
\textcolor{gray!55}{SegAnyMo$^{\dagger}$}
& \textcolor{gray!55}{0.78} & \textcolor{gray!55}{0.66} & \textcolor{gray!55}{0.77} & \textcolor{gray!55}{0.70}
& \textcolor{gray!55}{0.72} & \textcolor{gray!55}{0.50} & \textcolor{gray!55}{0.62} & \textcolor{gray!55}{0.55}
& \textcolor{gray!55}{0.77} & \textcolor{gray!55}{0.58} & \textcolor{gray!55}{0.66} & \textcolor{gray!55}{0.58} \\
SM4RT+HDBSCAN
& \textbf{0.84} & 0.54 & \underline{0.70} & 0.59
& \textbf{0.71} & 0.43 & \textbf{0.56} & 0.47
& \textbf{0.68} & 0.42 & \underline{0.53} & \underline{0.45} \\
MoSE3+HDBSCAN
& 0.39 & \underline{0.65} & 0.51 & \underline{0.69}
& 0.35 & \textbf{0.56} & \underline{0.49} & \textbf{0.66}
& 0.34 & \underline{0.53} & \underline{0.46} & \textbf{0.59} \\
\midrule
\multicolumn{13}{l}{\textit{Unsupervised}} \\
Progressive-X
& 0.65 & 0.44 & 0.59 & 0.49
& \underline{0.52} & 0.41 & 0.45 & 0.45
& \underline{0.53} & 0.34 & 0.40 & 0.35 \\
\rowcolor{tableblue}
\textbf{RICOLA (Ours)}
& \underline{0.73} & \textbf{0.69} & \textbf{0.72} & \textbf{0.77}
& 0.46 & \underline{0.49} & 0.46 & \underline{0.54}
& \textbf{0.68} & \textbf{0.54} & \textbf{0.58} & \textbf{0.59} \\
\bottomrule
\end{tabular}
\caption{\textbf{Moving rigid object discovery on Kubric with different trajectory sources.}
We evaluate using ground-truth 3D point tracks and estimated tracks from MoSE3 and TrackEverything.
$^{\dagger}$SegAnyMo is trained on our test set and is therefore excluded from the comparison.
Bold and underlined values indicate the best and second-best distinct scores among the compared methods, respectively; ties share the same marking.
Our method achieves the best or tied-best scores on three of four metrics with both GT and TrackEverything trajectories.}
\label{tab:kubric_main_table}
\end{table*}

\section{Experiments}
We evaluate on rigid object discovery from cluttered scenes (\cref{sec:exp_kubric}), rigid part discovery from articulated objects (\cref{sec:exp_itaco}), show in-the-wild qualitative results (\cref{sec:in_the_wild}), ablate key components (\cref{sec:ablation}), and demonstrate an application in object 6D pose tracking (\cref{sec:exp_ho3d}).

\subsection{Setup}
\noindent\textbf{Benchmarks.} We evaluate our method on rigid component discovery in Kubric~\cite{kubric} test set and iTACO~\cite{peng2026itaco}. Kubric contains 500 cluttered scenes with moving rigid objects. ITACO~\cite{peng2026itaco} is a synthetic articulated object dataset containing 11 categories of objects across 784 videos. For downstream application, we use 13 test videos from HO3D~\cite{ho3d} to evaluate 6D object pose tracking in real-world hand-object interaction videos.\\
\noindent\textbf{Baselines.}
We compare against two groups of methods: (a) \textit{supervised} methods, SegAnyMo~\cite{seganymo}, SM4RT~\cite{sm4rt}, and MoSE3~\cite{mose3}; and (b) \textit{unsupervised} methods, Progressive-X~\cite{barath2019progressive}. Here, supervised methods are those trained with task-specific annotations, such as moving object masks~\cite{seganymo} or SE(3) transform~\cite{mose3,sm4rt}. MotionBits~\cite{qian2026motionbits} is closest to our task setting, but we have to omit it as no public implementation was available at the time of evaluation. As these methods target different motion understanding tasks, we adapt their outputs to a common rigid-component discovery evaluation protocol for a consistent comparison.\\
\noindent\textbf{Evaluation Protocol.} We evaluate all methods on a common set of trajectories to ensure consistent comparison of rigid component assignments. We select query points visible in the first frame according to ground-truth annotations for estimating 3D point tracks. For SM4RT and MoSE3, we use HDBSCAN~\cite{mcinnes2017hdbscan} to obtain rigid components from their learned motion embeddings using the hyperparameters specified in their respective papers. For methods directly producing per-pixel component masks~\cite{seganymo,mose3,sm4rt}, we evaluate their predictions at the corresponding 2D points of 3D tracks. We select the best inlier threshold for Progressive-X on validation sets for each evaluation setting. Please see supplementary material for more details.\\
\noindent\textbf{Metrics.} We report Adjusted Rand Index (ARI) over the full video following object discovery evaluation protocols~\cite{slotcontrast}. ARI measures the similarity between two clusters corrected for random chance, where $1$ indicates perfect agreement and $0$ corresponds to random clustering. We also report mean Intersection over Union (mIoU), Recall (R.) and F1 at an IoU threshold of $0.5$ under one-to-one Hungarian matching.

\subsection{Moving Rigid Object Discovery}
\label{sec:exp_kubric}
\begin{figure*}
    \centering
    \footnotesize
    \begin{tabularx}{\linewidth}{
        @{}*{7}{>{\centering\arraybackslash}X}@{}
    }
        Input Video &
        SegAnyMo &
        MoSE3 &
        SM4RT & 
        Progressive-X &
        Ours &
        GT
    \end{tabularx}
   
    \includegraphics[width=\linewidth]{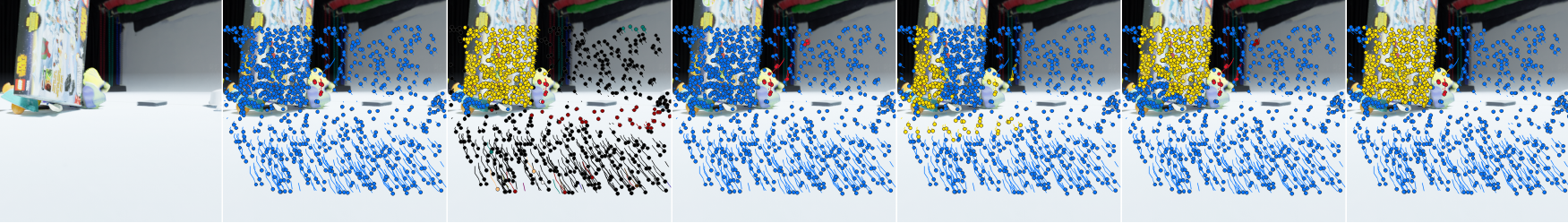}
     \begin{tabularx}{\linewidth}{
        @{}*{7}{>{\centering\arraybackslash}X}@{}
    }
         &
        \# Clusters=2 &
        \# Clusters=36 &
        \# Clusters=2 & 
        \# Clusters=2 &
        \# Clusters=4 &
        \# Clusters=3
    \end{tabularx}
    
    \includegraphics[width=\linewidth]{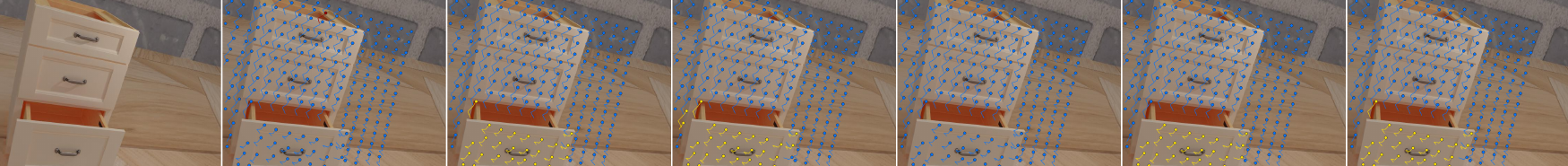}
    \begin{tabularx}{\linewidth}{
        @{}*{7}{>{\centering\arraybackslash}X}@{}
    }
         &
        \# Clusters=1 &
        \# Clusters=2 &
        \# Clusters=2 & 
        \# Clusters=1 &
        \# Clusters=2 &
        \# Clusters=2
    \end{tabularx}
    
    \vspace{3pt}
    
    \caption{
        \textbf{Qualitative comparison of rigid component discovery on Kubric (top) and iTACO (bottom).} Our method produces compact rigid component decompositions across diverse scenes. 
        } 
    \label{fig:qualitative}

\end{figure*}

We evaluate rigid object discovery on Kubric (MoVi-F) ~\cite{kubric} using three trajectory sources: ground-truth (GT), MoSE3~\cite{mose3}, and TrackEverything~\cite{jain2026trackeverything} (see \cref{tab:kubric_main_table}). 
Note that Kubric GT dynamic labels objects as moving if their initial placement is not on the floor, objects moved later, e.g., through collision, are thus labeled static. Since this penalizes the F1 score, we also report recall to assess the detection of annotated moving components only. SegAnyMo is trained on the test set with ground-truth motion labels; we show it in the table but exclude it from the ranking. Our self-supervised method substantially improves component-level segmentation over Progressive-X and remains competitive with supervised methods. SM4RT achieves the highest ARI but often undersegments rigid components, leading to low recall. MoSE3 produces meaningful SE(3) motion embeddings and achieves high mIoU and recall, but HDBSCAN often fragments static regions, resulting in low ARI (see ~\cref{fig:qualitative} top). In contrast, our method produces more coherent scene-level groupings. We note that our method tends to merge instances with similar motion, as we define the components from rigid motion only (Sec.~\ref{sec:problem}) without any spatial connectivity constraint, whereas the ground-truth annotation is based on instances.

\begin{table*}[t]
\centering
\small
\setlength{\tabcolsep}{4.5pt}
\begin{tabular}{l cccc cccc cccc}
\toprule
& \multicolumn{4}{c}{\textbf{GT}}
& \multicolumn{4}{c}{\textbf{MoSE3}}
& \multicolumn{4}{c}{\textbf{TrackEverything}} \\
\cmidrule(lr){2-5}
\cmidrule(lr){6-9}
\cmidrule(lr){10-13}
\textbf{Method}
& \textbf{ARI} $\uparrow$
& \textbf{mIoU} $\uparrow$
& \textbf{F1} $\uparrow$
& \textbf{R.} $\uparrow$
& \textbf{ARI} $\uparrow$
& \textbf{mIoU} $\uparrow$
& \textbf{F1} $\uparrow$
& \textbf{R.} $\uparrow$
& \textbf{ARI} $\uparrow$
& \textbf{mIoU} $\uparrow$
& \textbf{F1} $\uparrow$
& \textbf{R.} $\uparrow$\\
\midrule

\multicolumn{13}{l}{\textit{Supervised with Rigid Component Labels}}  \\
SegAnyMo
& 0.51 & 0.68 & 0.74 & 0.74
& 0.51 & 0.67 & 0.73 & 0.73
& 0.48 & 0.66 & 0.70 & 0.70 \\
SM4RT+HDBSCAN
& 0.58 & 0.72 & 0.78 & 0.76
& 0.55 & 0.70 & \underline{0.77} & 0.75
& 0.48 & 0.67 & 0.74 & 0.72 \\
MoSE3+HDBSCAN
& 0.70 & 0.85 & 0.70 & 0.91
& \textbf{0.72} & \textbf{0.81} & 0.71 & \textbf{0.89}
& \textbf{0.71} & \textbf{0.84} & \underline{0.70} & \textbf{0.87} \\
\midrule

\multicolumn{13}{l}{\textit{Unsupervised}} \\
Progressive-X
& \textbf{0.97} & \textbf{0.96} & \textbf{0.98} & \textbf{0.97}
& 0.15 & 0.50 & 0.70 & 0.58
& 0.43 & 0.63 & 0.61 & 0.70 \\
\rowcolor{tableblue}
\textbf{RICOLA (Ours)}
& \underline{0.90} & \underline{0.91} & \underline{0.96} & \underline{0.96}
& \underline{0.64} & \underline{0.75} & \textbf{0.79} & \underline{0.85}
& \underline{0.62} & \underline{0.73} & \textbf{0.84} & \underline{0.83} \\
\bottomrule
\end{tabular}
\caption{
\textbf{Articulated moving rigid-part discovery on iTACO.}
Our method outperforms unsupervised methods on real tracks and achieves competitive performance compared to the best supervised methods.}
\label{tab:itaco_main_table}
\end{table*}

\subsection{Moving Rigid Part Discovery}
\label{sec:exp_itaco}
We further evaluate rigid part discovery for articulated objects on iTACO~\cite{peng2026itaco} in~\cref{tab:itaco_main_table}. Progressive-X achieves the best performance with GT trajectories, since it is designed for geometric multi-model fitting.
However, its ARI drops with estimated tracks from MoSE3 and TrackEverything, indicating its sensitivity to tracking errors. SegAnyMo also struggles with articulated objects, potentially due to the introduced instance-level bias of SAM2~\cite{sam2}, which is prompted from predicted dynamic labels to generate dense masks (see \cref{fig:qualitative} bottom). In contrast, our method maintains competitive performance on GT tracks and achieves the best performance across all metrics among unsupervised methods on both estimated trajectory sources, approaching the performance of the best supervised methods. 

\subsection{Qualitative and in-the-wild results}
\label{sec:in_the_wild}
\begin{figure*}
    \centering
    \includegraphics[width=\linewidth]{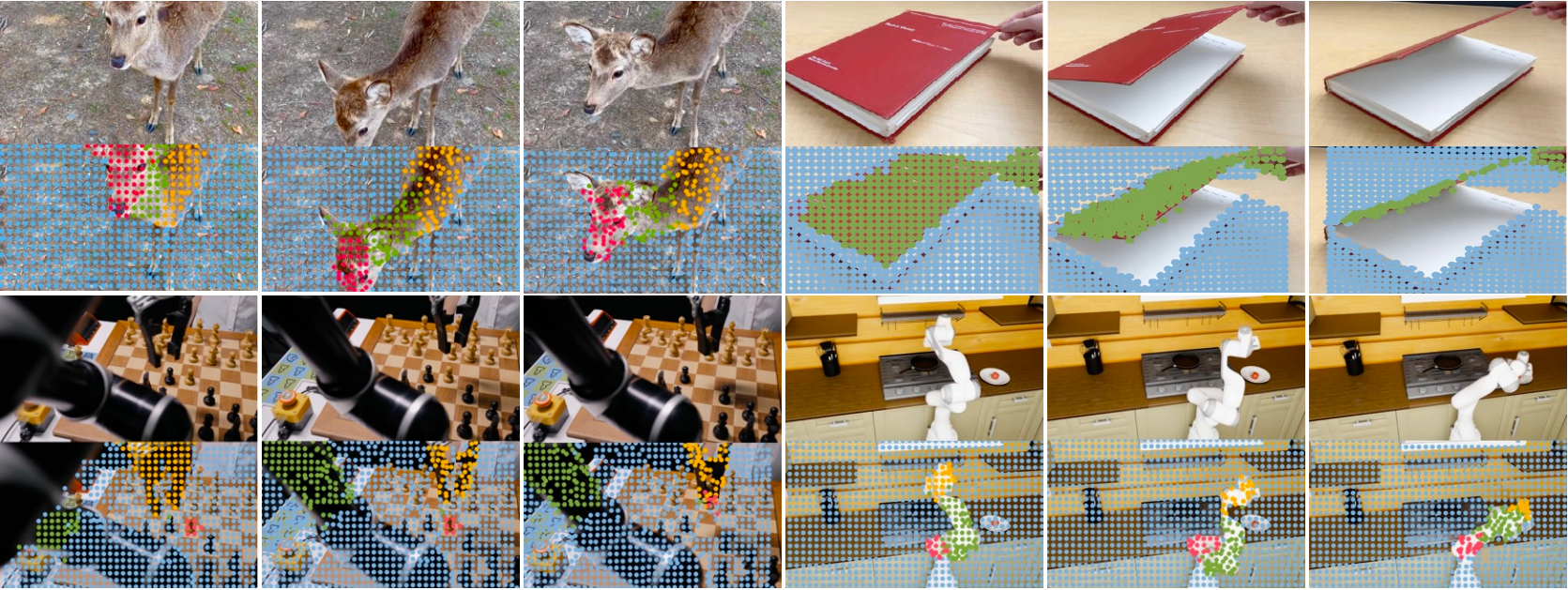}
    \caption{\textbf{Qualitative results on in-the-wild videos.} Our method learns to discover coherent rigid components trained on synthetic data only. We show the generalization to diverse real-world scenes, from articulated and slightly deformable objects (top) to cluttered interactions (bottom).}
    \label{fig:in_the_wild_results}
\end{figure*}

We provide in-the-wild qualitative results in ~\cref{fig:in_the_wild_results}. We use VGGT-$\Omega$~\cite{wang2026vggtomega} and TrackEverything~\cite{jain2026trackeverything} for camera pose estimation and 3D point tracking. Although our model is trained on synthetics data only, we show the generalization to casual videos across different domains, from articulated and slightly deforming objects to cluttered rigid-object interactions. In the hand-object interaction sequences, parts of the hand are grouped with the manipulated objects due to their similar motion. This shows our compact rigid component decompositions based on motion consistency rather than semantic object boundaries. If the hand lets go of the object at some point, as seen in robot chess example, our approach correctly separates them as distinct moving entities throughout the video. Please refer to the supplementary materials for more results.

\subsection{Ablations}
\label{sec:ablation}
We analyze the effects of our introduced components on Kubric (\cref{tab:ablation}). To save computational resources, all ablations are trained on a single GPU for one day.

\noindent\textbf{Architecture.}
We investigate the contributions of our architectural design in~\cref{tab:ablation} (a). 
Removing appearance or motion features substantially reduces F1 and Recall, which indicates missing moving rigid components. We replace slot attention with a standard transformer decoder, which also leads to a significant drop in all metrics. Removing explicit motion features has a smaller effect, as geometric motion information remains available through the training objectives. Together, these results highlight the importance of the fused motion representation and slot-based grouping to recover rigid components.
\begin{table}[t]
\centering
\small
\setlength{\tabcolsep}{5pt}
\resizebox{\columnwidth}{!}{
\begin{tabular}{lcccc}
\toprule
\textbf{Method} & \textbf{ARI} $\uparrow$ &  \textbf{mIoU} $\uparrow$ & \textbf{F1} $\uparrow$ & \textbf{R.} $\uparrow$\\

\midrule
\multicolumn{5}{l}{\textit{(a) Architecture}} \\
w/o Motion Feature
& +0.07 & +0.00 & -0.04 & -0.04 \\
w/o Appearance Feature
& +0.10 & -0.08 & -0.13 & -0.15 \\
w/o Slot Attention
& -0.03 & -0.01 & -0.12 & -0.03 \\
\midrule
\multicolumn{5}{l}{\textit{(b) Training Loss}} \\
w/o $\mathcal{L}_{feat}$
& +0.00 & -0.03 & -0.06 & -0.04 \\
w/o $\mathcal{L}_{red}$
& -0.35 & -0.06 & -0.24 & -0.12 \\
w/o $\mathcal{L}_{rigid}$ \& $\mathcal{L}_{red}$
& -0.43 & -0.18 & -0.34 & -0.29 \\
Direct slot count reg. & -0.06 & -0.10 & -0.12 & -0.12 \\
\midrule
\textbf{RICOLA (Ours, full)}
& \textbf{0.58}
& \textbf{0.53}
& \textbf{0.55}
& \textbf{0.59} \\
\bottomrule
\end{tabular}
}
\caption{
\textbf{Ablation study on Kubric with tracks from TrackEverything.}
We report performance differences relative to the full model (last row). w/o Slot Attention uses a standard transformer decoder for grouping.
}
\label{tab:ablation}
\end{table}

\noindent\textbf{Training objectives.}
We further study our self-supervised objectives in~\cref{tab:ablation} (b). Removing both geometric constraints ${\mathcal{L}_{rigid}}$ and $\mathcal{L}_{red}$ severely degrades grouping performance. Trained with $\mathcal{L}_{rigid}$ without $\mathcal{L}_{red}$ leads to unnecessary fragmentation and substantial performance drops. The complementarity of $\mathcal{L}_{red}$ prevents over segmentation. Feature reconstruction loss $\mathcal{L}_{feat}$ provides a complementary appearance signal, with a smaller but consistent benefit. We also evaluate direct slot count regularization (reg.) using similar slot attention and regularizer as AdaSlot~\cite{fan2024adaptive} in place of $\mathcal{L}_{red}$ , which results in degradation in all metrics. Overall, effective self-supervised component discovery requires both geometric consistency and constraints that encourage a compact and discriminative partition.

\noindent\textbf{Grouping Strategies.}
We compare our slot attention-based grouping against K-means and HDBSCAN in ~\cref{tab:merging_ablation}. Here, K-means and HDBSCAN  directly cluster trajectories to components based on our trajectory features. Slot attention consistently outperforms both clustering alternatives across Kubric and iTACO, even when K-means is provided with the ground-truth number of components. This indicates the importance of jointly learning trajectory representation and grouping mechanism.

\begin{table}[t]
\centering
\small
\setlength{\tabcolsep}{5pt}
\resizebox{\columnwidth}{!}{
\begin{tabular}{lcccccc}
\toprule
& \multicolumn{3}{c}{\textbf{Kubric}} & \multicolumn{3}{c}{\textbf{iTACO}} \\
\cmidrule(lr){2-4} \cmidrule(lr){5-7}
\textbf{Method} & \textbf{ARI} $\uparrow$ & \textbf{mIoU} $\uparrow$ & \textbf{F1} $\uparrow$ & \textbf{ARI} $\uparrow$ & \textbf{mIoU} $\uparrow$ & \textbf{F1} $\uparrow$\\
\midrule
K-means w. Oracle K
& 0.26 & 0.36 & 0.37
& 0.48 & 0.65 & 0.75\\
HDBSCAN 
& 0.17 & 0.45 & 0.29
& 0.55 & 0.69 & 0.71\\
\textbf{RICOLA (Slot Attention)}
& \textbf{0.68} & \textbf{0.54} & \textbf{0.58}
& \textbf{0.62} & \textbf{0.73} & \textbf{0.84} \\
\bottomrule
\end{tabular}
}
\caption{
\textbf{Ablation study of grouping strategies on Kubric and iTACO.} Our slot attention consistently achieves more accurate decompositions, even without oracle component counts.}
\label{tab:merging_ablation}
\end{table}

\section{Downstream Application}
\label{sec:exp_ho3d}
\textbf{SE(3) Pose Tracking.} 
We show our method can support the downstream task in object-level SE(3) pose tracking on HO3D~\cite{ho3d} (\cref{tab:pose_tracking}). Following the object-level evaluation protocol of MoSE3, we match predicted and ground-truth object masks under the one-to-one evaluation protocol, then run Weighted Umeyama to solve object poses. The results show that our method provides useful object clusters for object pose tracking without task-specific training. 


\begin{table}[t]
\centering
\resizebox{\columnwidth}{!}{
\begin{tabular}{lcccc}
\toprule
\textbf{Method} & \textbf{RRE} $\downarrow$ &
\textbf{ADD} $\downarrow$ &
\textbf{AUC}$_{\mathrm{ADD}}$ $\uparrow$ &
\textbf{mIoU} $\uparrow$ \\
\midrule
GT Object Mask & 36.99 & 2.56 & 75.40 & 1.00 \\
MoSE3 + HDBSCAN & 36.08 & 2.54 & 75.70 & 0.93 \\
Progressive-X & 38.22 & 4.50 & 61.80 & 0.23 \\
\midrule
\textbf{RICOLA (Ours)} & 41.20 & 2.81 & 72.20 & 0.77 \\
\bottomrule
\end{tabular}
}
\caption{
\textbf{Object-level SE(3) estimation on the HO3D dataset.} MoSE3 is used for point tracking across all methods. Our method can applied for pose tracking tasks without task-specific supervision.
}
\label{tab:pose_tracking}
\end{table}


\begin{figure}
    \centering
    \includegraphics[width=\linewidth]{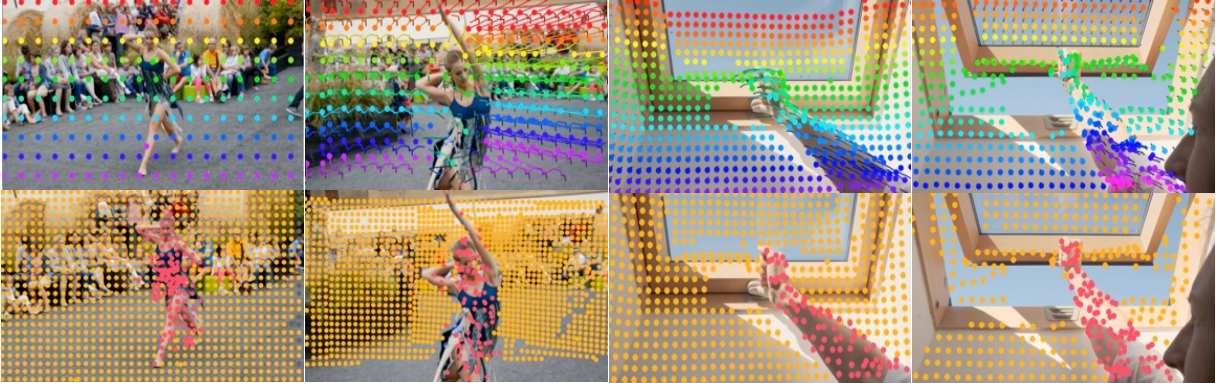}
    \caption{\textbf{Failure Cases.} Left: complex non-rigid motion; Right: tracking failure with missing tracks at the opened window. 
    }
    \label{fig:failures}
\end{figure}

\section{Limitations}
Our method has several limitations. First, we use a fixed number of \(K\) slots as an upper bound on component capacity. Therefore, scaling up to a larger model requires retraining. Second, our grouping directly depends on the quality of the input 3D trajectories. While we show robustness when run with recent trackers, large errors can break our grouping (\cref{fig:failures} right). Third, our method struggles with complex non-rigid motion (\cref{fig:failures} left). Finally, we do not consider explicit spatial connectivity constraints, therefore disconnected objects with a similar motion would be grouped together.

\section{Conclusion}
We presented RICOLA, a self-supervised framework to discover rigid components from 3D point trajectories without requiring rigid component annotations. By jointly learning trajectory representations and rigid component assignments, our method produces consistent groupings through geometric rigidity and redundancy constraints. Experiments on rigid and articulated scenes demonstrate its effectiveness across different trajectory sources. We hope this work provides a foundation for learning structured scene representations from motion, enabling further progress in dynamic scene understanding.

\section{LLM Usage Statement}
We use LLMs in this work to assist with both the writing and the research. ChatGPT is used to proofread and review the manuscript, search for related work, generate sketches of figures, and help formulate research hypotheses. We also use Claude to assist with coding and implementation. Finally, we use Deep Review to review the manuscript. The authors take full responsibility for the implementation, experimental results, cited literature, and final manuscript.



{
    \small
    \bibliographystyle{ieeenat_fullname}
    \bibliography{main}
}

\clearpage
\setcounter{page}{1}
\maketitlesupplementary

\section{Training Details}
\subsection{Parameters}
We train our model with batch size 36 on each GPU. We set the feature dimension $D=128$, $K=10$, $\delta=2e^{-5}, \tau=\delta /4 $ for our experiments. We use AdamW optimizer with a cosine learning rate schedule. We use a warm up for the $\lambda_{red}$ for the first 2k step from 0 to 0.1. For the first stage, we use $1e^{-4}$ as the learning rate with 2000 warmup steps. Then for the second stage, we train on a mixed training data, including Kubric~\cite{kubric}, ArtKubric~\cite{mose3}, Synthverse~\cite{synthverse} (articulation, embodied, flowers, animals), sampling each dataset with equal probability.

\subsection{Data Augmentation}
During training, we uniformly sample the temporal stride from $\{1,2,3,4\}$ and the clip length from $\{8,16,20,24\}$. We use photometric, 2D tracks and depth augmentations to improve the model generalization. We use various photometric augmentations including eraser, random zoom, flips, crops, scale jitter, color jitter, and gaussian blur. We define 3D trajectories in the first-frame camera coordinate system by applying the coordinate transformation. Then, we use track augmentation to simulate tracking errors through 2D localization offsets, temporal drift, and depth perturbations. 

\subsection{Supervision Setting} Our method does not require ground-truth rigid-component labels or SE(3) poses. Instead, we use geometric rigidity constraints as the self-supervision signal. During training, ground-truth trajectories from synthetic datasets~\cite{kubric,mose3,synthverse} are used to compute the training objectives, while the network takes noise-augmented trajectories as inputs to improve robustness to tracking errors. At inference, only estimated 3D trajectories and RGB videos are required.

\section{Evaluation Metrics}
We evaluate all methods on point trajectories to ensure a consistent comparison across different prediction representations. We evaluate segmentation quality using Adjusted Rand Index (ARI) and mean Intersection-over-Union (mIoU) over time to measure both the segmentation quality and temporal consistency. We compute video ARI over all valid trajectory observations and video mIoU using one-to-one Hungarian matching based on IoU aggregated across frames. For Kubric~\cite{kubric}, we use the ground-truth dynamic instance masks. For iTACO~\cite{peng2026itaco}, we use the labels derived from ground-truth motion exceeding $1^\circ$ rotation or $1\mathrm{cm}$ translation. Component precision, recall, and F1 are computed from trajectory-level labels with an IoU threshold of 0.5. For methods producing per-frame segmentation masks, predictions are sampled at the corresponding trajectory positions, with majority voting used to obtain trajectory-level labels. 

\section{Additional Ablation Studies}
We ablate other important components of our method. All ablations are trained on Kubric for one day and evaluated on the tracks predicted by TrackEverything~\cite{jain2026trackeverything}.

\subsection{Number of Learnable Slot Queries.}
\cref{tab:ablation_K} reports the results with different number of learnable queries for slot attention. Kubric contains 4 to 6 rigid components, such that limiting the model to five slots may prevent it from discovering all components, leading to reduced recall. Our model maintains highly similar performance across different slot budgets, even when increasing $K$ from 10 to 30. Although using more slots leads to a slight reduction in F1, the overall results demonstrate that our framework is robust to variations in the slot budget, without severe degradation in grouping quality. This highlights the efficacy of our proposed slot redundancy loss where over-segmentation from up to 30 slots can be recovered.
We adopt $K=10$ as the default configuration for its favorable balance between performance and model complexity. 

\begin{table}[t]
\centering
\small
\setlength{\tabcolsep}{5pt}
\begin{tabular}{lccccc}
\toprule
\textbf{K} & \textbf{ARI} $\uparrow$ &  \textbf{mIoU} $\uparrow$ & \textbf{F1} $\uparrow$ & \textbf{R.} $\uparrow$ \\
\midrule
5 & 0.55 & 0.48 & 0.54 & 0.53 \\
10 (default) &{0.58} &{0.53} &{0.55} &{0.59}\\
15 & 0.55 & 0.51 & 0.54 & 0.57 \\
20 & 0.54 & 0.51 & 0.52 & 0.57 \\
30 & 0.58 & 0.49 & 0.49 & 0.53 \\
\bottomrule
\end{tabular}
\caption{
\textbf{Ablation study on the Number of Learnt Slot Queries on Kubric.} We use $10$ as the default in our main paper.
}
\label{tab:ablation_K}
\end{table}

\subsection{Sensitivity of Parameters for Redundancy Loss} 
\cref{tab:ablation_delta,tab:ablation_tau} ablate the threshold $\delta$ and temperature $\tau$ in the redundancy loss $\mathcal{L}_{red}$ on Kubric. Varying $\delta$ across a wide range yields relatively stable segmentation performance. Notably, increasing $\delta$ to $3\times10^{-4}$ improves ARI to 0.61 while maintaining the same mIoU and F1 as the default setting. Similarly, the model performs consistently for $\tau\in\{\delta/2,\delta/4,\delta/6\}$, with only minor variations across metrics. However, further reducing $\tau$ to $\delta/8$ degrades performance, suggesting that an excessively small temperature can negatively affect component grouping. Overall, these results demonstrate that our redundancy loss is robust to hyperparameter variations and not sensitive to careful tuning. We adopt $\delta=2\times10^{-5}$ and $\tau=\delta/4$ as the default configuration, achieving a favorable balance across the evaluation metrics.

\begin{table}[t]
\centering
\small
\setlength{\tabcolsep}{5pt}
\begin{tabular}{lccccc}
\toprule
\textbf{$\delta$} & \textbf{ARI} $\uparrow$ 
&  \textbf{mIoU} $\uparrow$ 
& \textbf{F1} $\uparrow$ 
& \textbf{R.} $\uparrow$\\
\midrule
$1e-5$ & 0.55 & 0.53 & 0.53 & 0.59 \\
$2e-5$ (default)
&{0.58} &{0.53} &{0.55} &{0.59} \\
$5e-5$ & 0.54 & 0.52 & 0.52 & 0.57 \\
$1e-4$ & 0.54 & 0.53 & 0.49 & 0.59\\
$3e-4$ & 0.61 & 0.53 & 0.55 & 0.59 \\
\bottomrule
\end{tabular}
\caption{
\textbf{Ablation study on the $\delta$.} We use $2e-5$ as the default in our main paper.
}
\label{tab:ablation_delta}
\end{table}

\begin{table}[t]
\centering
\small
\setlength{\tabcolsep}{5pt}
\begin{tabular}{lccccc}
\toprule
\textbf{$\tau$} & \textbf{ARI} $\uparrow$ 
&  \textbf{mIoU} $\uparrow$ 
& \textbf{F1} $\uparrow$ 
& \textbf{R.} $\uparrow$\\
\midrule
$\delta / 2$ & 0.55 & 0.54 & 0.53 & 0.58 \\
$\delta / 4$ (default) &{0.58} &{0.53} &{0.55} &{0.59} \\
$\delta / 6$ & 0.58 & 0.54 & 0.53 & 0.58\\
$\delta / 8$ & 0.48 & 0.51 & 0.49 & 0.56\\
\bottomrule
\end{tabular}
\caption{
\textbf{Ablation study on the temperature $\tau$ in redundancy loss.} We use $\delta / 4$ as the default in our main paper.
}
\label{tab:ablation_tau}
\end{table}


\subsection{Visualizing Different Grouping Strategies.}
\begin{figure}
    \centering
    \footnotesize
    \includegraphics[width=\linewidth]{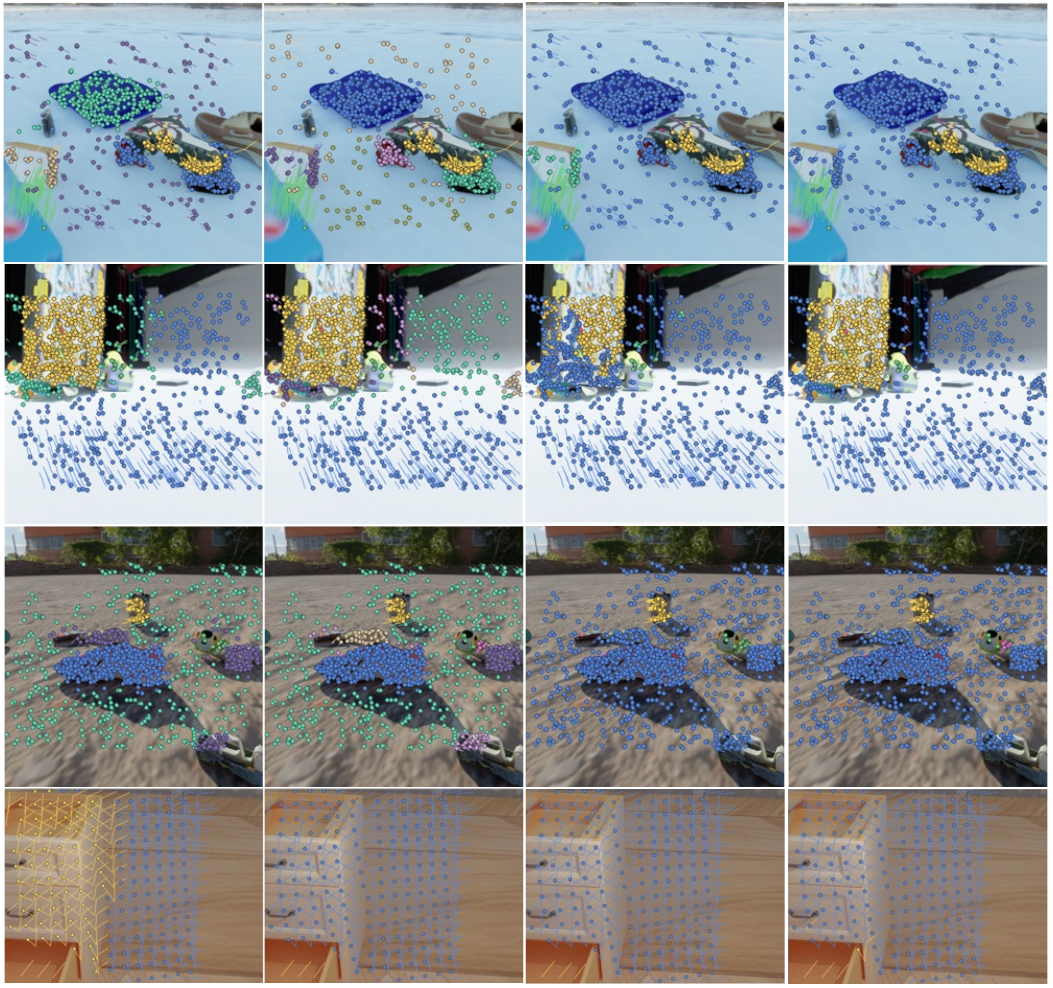}
    \begin{tabularx}{\linewidth}{
        @{}*{4}{>{\centering\arraybackslash}X}@{}
    }
        K-means &
        HDBSCAN &
        Ours &
        GT
    \end{tabularx}
    \caption{\textbf{Qualitative results for different grouping strategies.} Our slot attention consistently achieves more accurate decompositions compared to K-means and HDBSCAN.}
    \label{fig:vis_merging_ablation}
 \end{figure}

\cref{fig:vis_merging_ablation} presents the qualitative results corresponding to~\cref{tab:merging_ablation} in the main paper. The results suggest that learning the grouping using slot attention is more effective than using clustering algorithms to discover rigid components.

\section{Inference Speed}
\begin{figure}
    \centering
    \includegraphics[width=\linewidth]{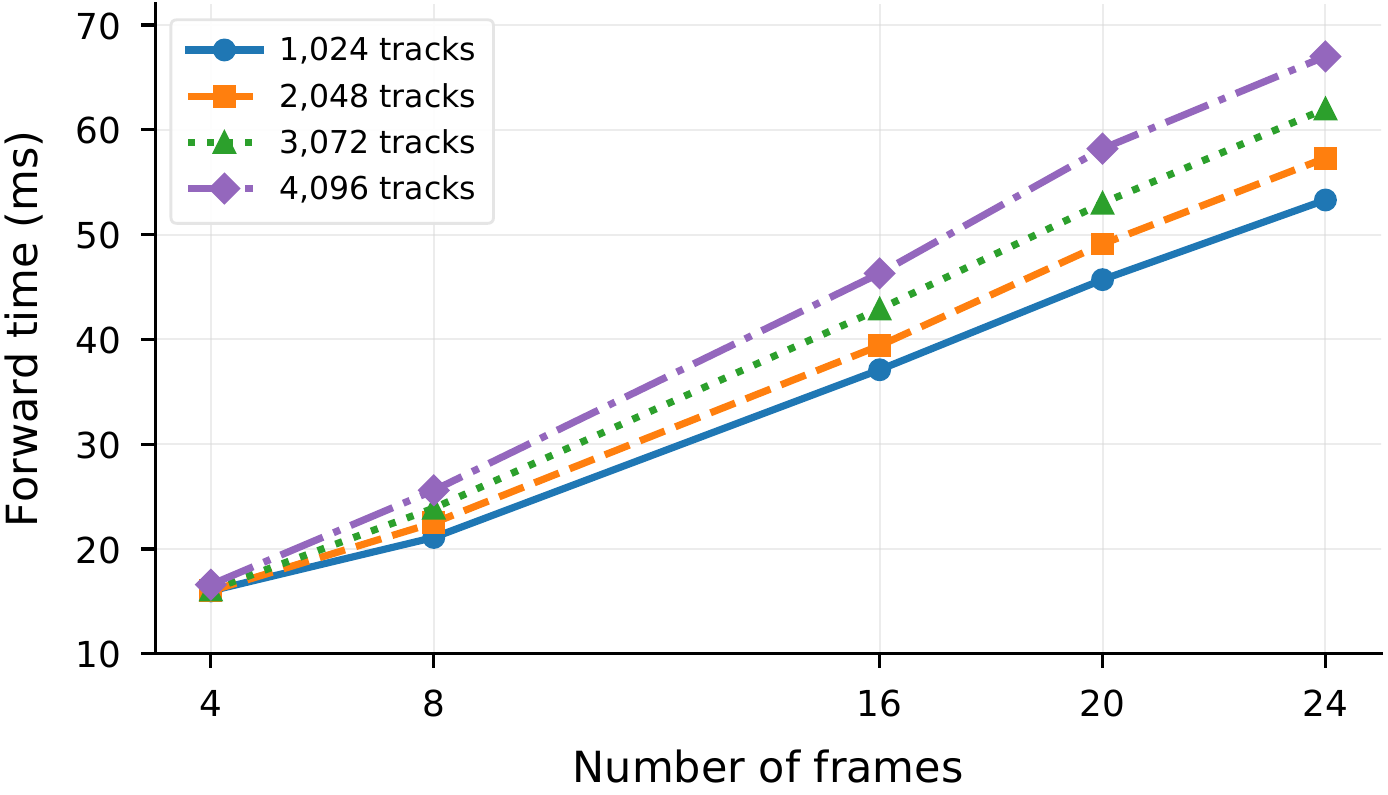}
    \caption{Per forward pass time for varying numbers of frames and trajectories in ms. }
    \label{fig:inference_speed}
\end{figure}

\cref{fig:inference_speed} presents the time per forward pass of our model for varying input sizes. Our runtime scales with sequence length and the number of trajectories. However, it the inference time remains below 70\,ms for all settings and achieves a throughput of 358–450 frames per second for 24-frame clips. The reported runtime excludes the upstream 3D point tracking. 

\section{Progressive-X Parameter Sweep}
\begin{figure*}
    \centering
    \includegraphics[width=\linewidth]{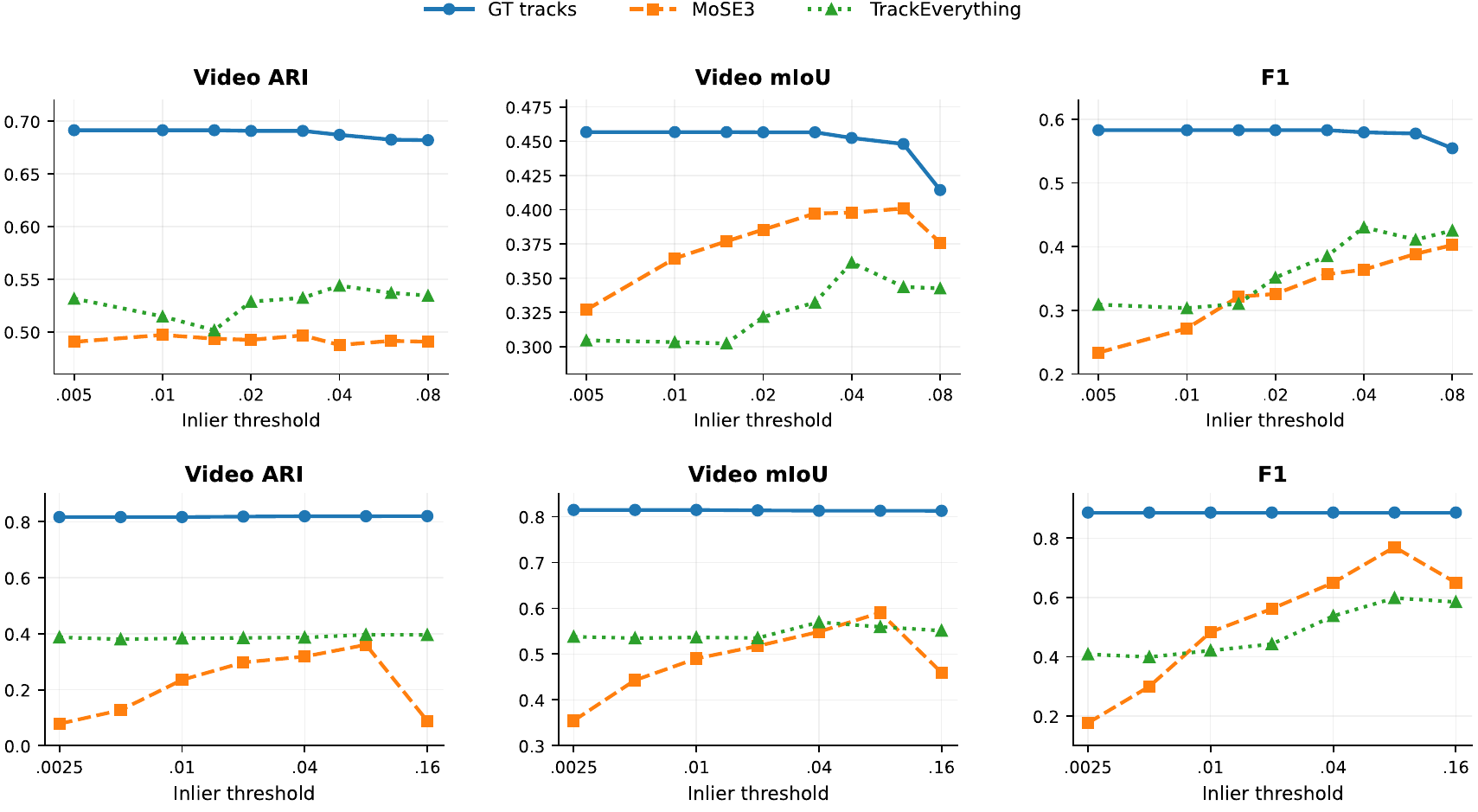}

    \caption{\textbf{Progressive-X inlier threshold sensitivity on Kubric (top) and iTACO (bottom) validation sets.} Performance remains relatively stable across thresholds with ground-truth trajectories but is more sensitive to threshold selection with estimated trajectories.}
    \label{fig:progx_sweep}
\end{figure*}

We select the best inlier threshold for Progressive-X~\cite{barath2019progressive} on 20-video validation sets based on mIoU metrics. Results in~\cref{fig:progx_sweep} show that performance with ground-truth tracks is relatively stable, however, estimated tracks show greater sensitivity. In contrast, our method does not require careful parameter tuning. We compare against the best version of Progressive-X for a favorable comparison.

\section{Detailed Results}
For completeness, we additionally report precision for \cref{tab:kubric_main_table,tab:itaco_main_table}. However, Kubric has missing moving object annotations for objects that become dynamic through interactions (e.g., collisions). Therefore, correctly discovered moving components can be counted as false positives, leading to an underestimation of precision.

\begin{table*}[t]
\centering
\small
\setlength{\tabcolsep}{4.5pt}
\resizebox{\textwidth}{!}{
\begin{tabular}{l*{15}{c}}
\toprule
& \multicolumn{5}{c}{\textbf{GT}}
& \multicolumn{5}{c}{\textbf{MoSE3}}
& \multicolumn{5}{c}{\textbf{TrackEverything}} \\
\cmidrule(lr){2-6}
\cmidrule(lr){7-11}
\cmidrule(lr){12-16}
\textbf{Method}
& \textbf{ARI} $\uparrow$
& \textbf{mIoU} $\uparrow$
& \textbf{F1} $\uparrow$
& \textbf{R.} $\uparrow$
& \textbf{P.} $\uparrow$
& \textbf{ARI} $\uparrow$
& \textbf{mIoU} $\uparrow$
& \textbf{F1} $\uparrow$
& \textbf{R.} $\uparrow$
& \textbf{P.} $\uparrow$
& \textbf{ARI} $\uparrow$
& \textbf{mIoU} $\uparrow$
& \textbf{F1} $\uparrow$
& \textbf{R.} $\uparrow$
& \textbf{P.} $\uparrow$
\\
\midrule
\multicolumn{16}{l}{\textit{Supervised with Rigid Component Labels}} \\
\textcolor{gray!55}{SegAnyMo$^{\dagger}$}
& \textcolor{gray!55}{0.78} & \textcolor{gray!55}{0.66} & \textcolor{gray!55}{0.77} & \textcolor{gray!55}{0.70} & \textcolor{gray!55}{0.88}
& \textcolor{gray!55}{0.72} & \textcolor{gray!55}{0.50} & \textcolor{gray!55}{0.62} & \textcolor{gray!55}{0.55} & \textcolor{gray!55}{0.76}
& \textcolor{gray!55}{0.77} & \textcolor{gray!55}{0.58} & \textcolor{gray!55}{0.66} & \textcolor{gray!55}{0.58} & \textcolor{gray!55}{0.80} \\
SM4RT+HDBSCAN
& \textbf{0.84} & 0.54 & \underline{0.70} & 0.59 & \textbf{0.90}
& \textbf{0.71} & 0.43 & \textbf{0.56} & 0.47 & \textbf{0.73}
& \textbf{0.68} & 0.42 & \underline{0.53} & \underline{0.45} & \textbf{0.69} \\
MoSE3+HDBSCAN
& 0.39 & \underline{0.65} & 0.51 & \underline{0.69} & 0.43
& 0.35 & \textbf{0.56} & \underline{0.49} & \textbf{0.66} & 0.45
& 0.34 & \underline{0.53} & \underline{0.46} & \textbf{0.59} & 0.41 \\
\midrule
\multicolumn{16}{l}{\textit{Unsupervised}} \\
Progressive-X
& 0.65 & 0.44 & 0.59 & 0.49 & \underline{0.80}
& \underline{0.52} & 0.41 & 0.45 & 0.45 & \underline{0.51}
& \underline{0.53} & 0.34 & 0.40 & 0.35 & 0.50 \\
\rowcolor{tableblue}
\textbf{RICOLA (Ours)}
& \underline{0.73} & \textbf{0.69} & \textbf{0.72} & \textbf{0.77} & 0.69
& 0.46 & \underline{0.49} & 0.46 & \underline{0.54} & 0.42
& \textbf{0.68} & \textbf{0.54} & \textbf{0.58} & \textbf{0.59} & \underline{0.59} \\
\bottomrule
\end{tabular}
}
\caption{\textbf{Moving rigid object discovery on Kubric with different trajectory sources.} Due to incomplete motion annotations in Kubric, valid moving components may be counted as false positives, leading to underestimated precision. }
\label{tab:kubric_full_table}
\end{table*}

\begin{table*}[t]
\centering
\small
\setlength{\tabcolsep}{4.5pt}
\resizebox{\textwidth}{!}{
\begin{tabular}{l ccccc ccccc ccccc}
\toprule
& \multicolumn{5}{c}{\textbf{GT}}
& \multicolumn{5}{c}{\textbf{MoSE3}}
& \multicolumn{5}{c}{\textbf{TrackEverything}} \\
\cmidrule(lr){2-6}
\cmidrule(lr){7-11}
\cmidrule(lr){12-16}
\textbf{Method}
& \textbf{ARI} $\uparrow$
& \textbf{mIoU} $\uparrow$
& \textbf{F1} $\uparrow$
& \textbf{R.} $\uparrow$
& \textbf{P.} $\uparrow$
& \textbf{ARI} $\uparrow$
& \textbf{mIoU} $\uparrow$
& \textbf{F1} $\uparrow$
& \textbf{R.} $\uparrow$
& \textbf{P.} $\uparrow$
& \textbf{ARI} $\uparrow$
& \textbf{mIoU} $\uparrow$
& \textbf{F1} $\uparrow$
& \textbf{R.} $\uparrow$
& \textbf{P.} $\uparrow$
\\
\midrule

\multicolumn{13}{l}{\textit{Supervised with Rigid Component Labels}}  \\
SegAnyMo
& 0.51 & 0.68 & 0.74 & 0.74 & 0.88
& 0.51 & 0.67 & 0.73 & 0.73 & \underline{0.76}
& 0.48 & 0.66 & 0.70 & 0.70 & \underline{0.80} \\
SM4RT+HDBSCAN
& 0.58 & 0.72 & 0.78 & 0.76 & 0.89 
& 0.55 & 0.70 & \underline{0.77} & 0.75 & \textbf{0.89}
& 0.48 & 0.67 & 0.74 & 0.72 & 0.79\\
MoSE3+HDBSCAN
& 0.70 & 0.85 & 0.70 & 0.91 & 0.60
& \textbf{0.72} & \textbf{0.81} & 0.71 & \textbf{0.89} & 0.61
& \textbf{0.71} & \textbf{0.84} & \underline{0.70} & \textbf{0.87} & 0.62 \\
\midrule

\multicolumn{13}{l}{\textit{Unsupervised}} \\
Progressive-X
& \textbf{0.97} & \textbf{0.96} & \textbf{0.98} & \textbf{0.97} & \textbf{0.99}
& 0.15 & 0.50 & 0.70 & 0.58 & 0.96 
& 0.43 & 0.63 & 0.61 & 0.70 & 0.58\\
\rowcolor{tableblue}
\textbf{RICOLA (Ours)}
& \underline{0.90} & \underline{0.91} & \underline{0.96} & \underline{0.96} & \underline{0.97}
& \underline{0.64} & \underline{0.75} & \textbf{0.79} & \underline{0.85} & 0.67 
& \underline{0.62} & \underline{0.73} & \textbf{0.84} & \underline{0.83} & \textbf{0.87}\\
\bottomrule
\end{tabular}
}
\caption{
\textbf{Articulated moving rigid-part discovery on iTACO.}
Our method outperforms unsupervised methods on GT tracks and achieves competitive performance compared to the best supervised methods.}
\label{tab:itaco_full_table}
\end{table*}


\end{document}